\documentclass[10pt,twocolumn,letterpaper]{article}

\usepackage{cvpr}
\usepackage{times}
\usepackage{epsfig}
\usepackage{graphicx}
\usepackage{amsmath}
\usepackage{amssymb}
\usepackage{soul}
\usepackage{subfigure}
\usepackage{caption,booktabs,multirow,comment}
\usepackage[square,sort,comma,numbers]{natbib}

\usepackage[pagebackref=true,breaklinks=true,letterpaper=true,colorlinks,bookmarks=false]{hyperref}

 \cvprfinalcopy 

\def\cvprPaperID{539} 

\ifcvprfinal\fi
\begin{document}

\title{DimensionApp : android app to estimate object dimensions}

\author{
Suriya Singh\\
CVIT, IIIT Hyderabad
\and
Vijay Kumar\\
CVIT, IIIT Hyderabad
\and 
Prof. C.V. Jawahar\\
Course Instructor\\
Computer Vision, 2014
}

\maketitle
\begin{abstract}
In this project, we develop an android app that uses computer vision techniques to estimate an object dimension present in field of view. The app while having compact size, is accurate upto +/- 5 mm and robust towards touch inputs. We use single-view metrology to compute accurate measurement. Unlike previous approaches, our technique does not rely on line detection and can be generalize to any object shape easily.
\end{abstract}

\section{Introduction}
Smartphones have now replaced many measuring instruments and devices specially the ones that we use in our daily life. Examples are alarm clock, calendar, calculator, maps etc. With the help of computer vision, we also have apps that can monitor persons health and estimate blood pressure and heart beat rate accurately. Further, there are apps that can help visually impaired to read, identify currency or navigate through environment. This has been possible due to the fact that modern smartphones are packed with various sensors and increase in processing capabilities. Availability of affordable smartphones have put these apps to frequent use by more than 1 billion people everyday. One similar application could be to use smartphone to estimate dimensions of various objects, sketches and drawings. It can be used as an accurate and reliable measuring device and specifically to estimate 3D affine measurement from single view. 

In order to estimate objects dimensions we use single-view metrology. The single-view metrology method makes use of planes and parallel lines in the scene to extract the physical dimensions of structures from a single perspective view, given minimal prior knowledge about the scene. The prior knowledge usually involves the recognition of vanishing points of a reference plane, and a vanishing point of a direction not parallel to the reference plane. As a result, this method is especially suitable for scene structures that involve parallel lines, which could often be found in man-made structures, such as architectures and geometric elements. Concepts of the single-view metrology method were described in a number of papers  \cite{Reid, Kim} and were later generalized by A. Criminisi et al.  \cite{Criminisi}. To increase usability and reliability of such system, we will use a predefined reference object which will ensure we have enough information about the scene.

\begin{figure}[t]
    \centering
   \includegraphics[width=0.9\linewidth]{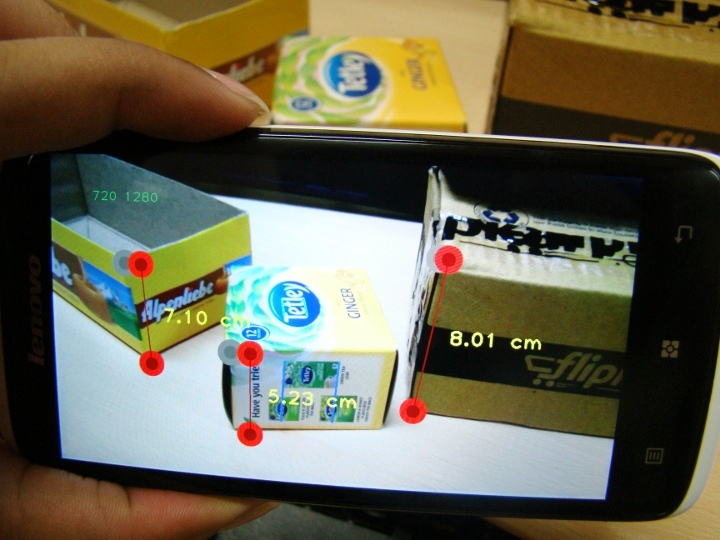}
   \caption{A typical use-case for DimensionsApp.}
    \label{fig:use_case}
\end{figure}

\section{Problem Statement}
To estimate dimensions of objects given a reference object in the scene.
 A typical use-case of the app is shown in Figure 1.
  In this project we consider measuring only one dimension at a time (height of boxes in Figure 1).
   Accuracy of measurement is naturally the first concern for such apps followed by speed. The app should be accurate, fast and robust while taking less memory and storage.
    Since we focus on usability of the app, it should be easy to use and capable of handling inaccurate input points due to touch interface.
    
\section{Challenges and Design}
Working on a mobile platform brings with it a number of unique challenges that need to be taken care of.
 Primarily, the restrictions are in the memory, the application size, and the processing time. 
Single-view metrology method can provide information about length only in reference direction.
Given the casual nature of everyday use and limitation of physical screen, we need to align inaccurate input points to the reference direction.
 Since projective transformation is involved while capturing image, adjusting slope of the line is not the solution.
  Further, since the app is to be used in all kinds of environment, finding parallel lines automatically in the scene is difficult.
   Moreover, parallel lines estimated this way are far from being accurate. Inaccuracy results in huge error in final estimate.

\section{Solution - Single view metrology}

\subsection{Finding Parallel Lines in the Scene}
As previously mentioned, automatically detecting parallel lines in the scene is a difficult task, and results are also not reliable. To overcome this, 
we find parallel lines with the help of reference object. To estimate the planar transformation of surfaces we compute homography between the scene and image of each side of the 
reference object (Figure 2.a). We used SIFT as a feature to find correspondences (Figure 2.b), however, ORB-FREAK also works fine with some restriction like size of reference object and how skewed is it in the scene.
Homography matrix $H$ is computed using RANSAC implementation of OpenCV. Using $H$, we get the predefined parallel lines in the scene.

\subsection{Finding Vanishing Line and Vanishing Point }
Using set of parallel lines obtained previously, we can easily determine vanishing line of reference plane and vanishing point in reference direction (Figure 2.c and 2.d).
Vanishing line \textbf{l} is computed by using 2 sets of parallel lines pair while vanishing point \textbf{v} is computed by using one set of parallel lines pair.

\subsection{Align User's Input to Reference Direction}
Due to nature of use of touch screens, input points  \textbf{b}$_x$ and \textbf{t}$_x$ (Figure 2.e) may deviate from reference direction. This deviation can cause huge error in 
further steps. To overcome this, we align user input points to reference direction before computing metrics. After user has provided input points \textbf{b}$_x$ and \textbf{t}$_x'$,
project \textbf{t}$_x'$ on line $\textbf{v} \times \textbf{b}_x$ to get point \textbf{t}$_x$. It can be easily verified that \textbf{b}$_x$ and \textbf{t}$_x$ are points in reference direction. Or that 
the line joining \textbf{b}$_x$ and \textbf{t}$_x$ also passes through \textbf{v}.

\subsection{Computing Metric Factor and Dimension}
The heights of scene elements could be estimated based on vanishing points and a known reference height found in 
the scene. See Fig.2(e) for an illustration of the method. Here \textbf{l} is the horizontal vanishing line, \textbf{v} is the vertical vanishing 
point, and $Z_r$ is the reference height. \textbf{b}$_r$ and \textbf{t}$_r$ are the bottom and top of the reference height, respectively. The metric factor $\alpha$ 
can then be found by the following equation:
\begin{equation}
\alpha Z_r = - \frac{\lVert \textbf{b}_r \times \textbf{t}_r \rVert}{\left( \textbf{l}.\textbf{b}_r \right)\lVert \textbf{v} \times \textbf{t}_r \rVert}
\end{equation}

For a given object on the reference plane, its height $Z_x$ can then be found by the following equation, where \textbf{b}$_x$ and \textbf{t}$_x$ are 
the bottom and the top of the object, respectively:
\begin{equation}
Z_x = - \frac{\lVert \textbf{b}_x \times \textbf{t}_x \rVert}{\alpha \left( \textbf{l}.\textbf{b}_x \right)\lVert \textbf{v} \times \textbf{t}_x \rVert}
\end{equation}

\begin{figure*}[H]
\centering 
\begin{subfigure}[]{\includegraphics[width=0.2\linewidth ]{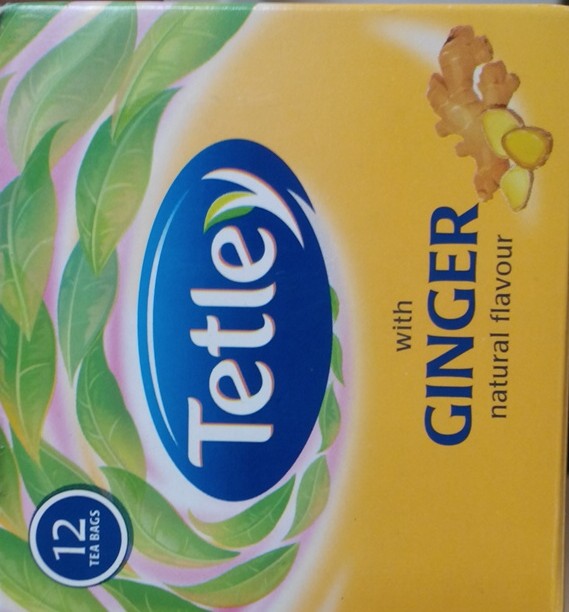}}
 \end{subfigure}
 \begin{subfigure}[]{\includegraphics[width=0.2\linewidth ]{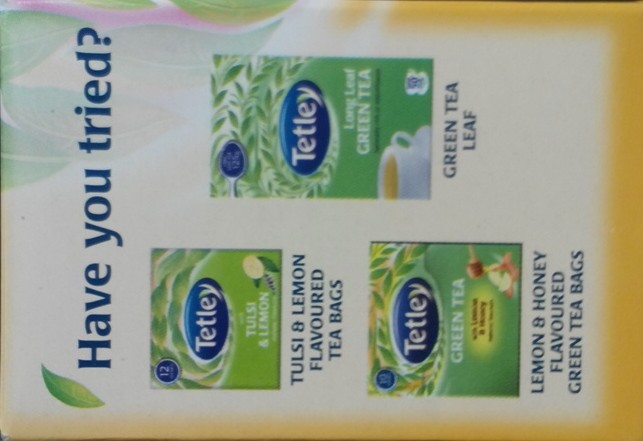}}
 \end{subfigure}
 \begin{subfigure}[]{\includegraphics[width=0.4\linewidth ]{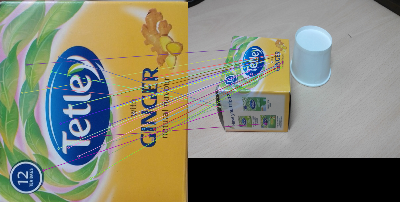}}\\
 \end{subfigure}
 \begin{subfigure}[]{\includegraphics[width=0.2\linewidth ]{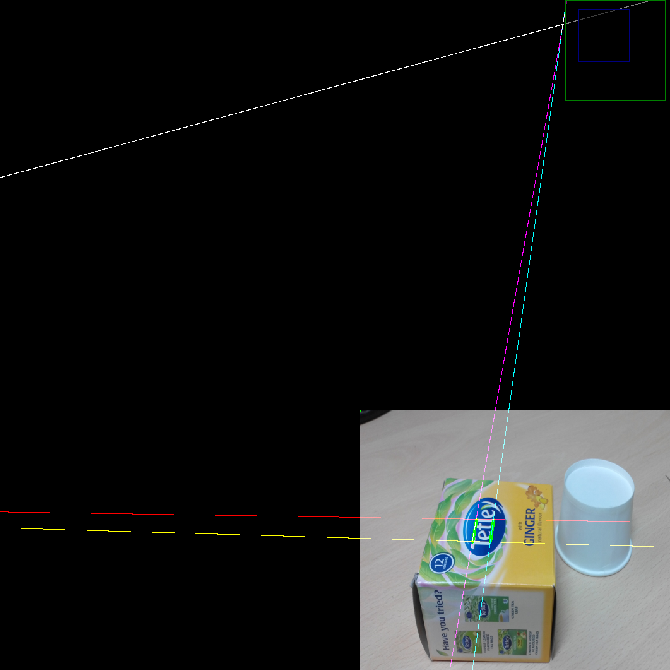}}
 \end{subfigure}
 \begin{subfigure}[]{\includegraphics[width=0.2\linewidth ]{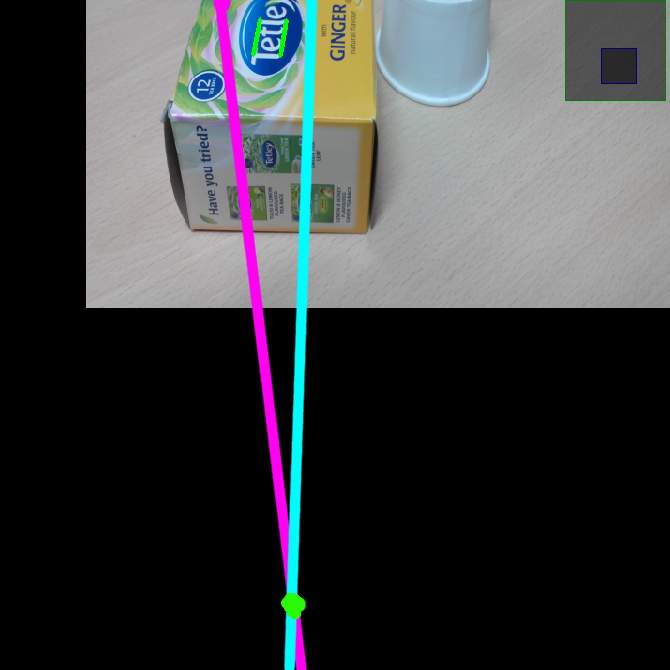}}
 \end{subfigure}
  \begin{subfigure}[]{\includegraphics[width=0.5\linewidth ]{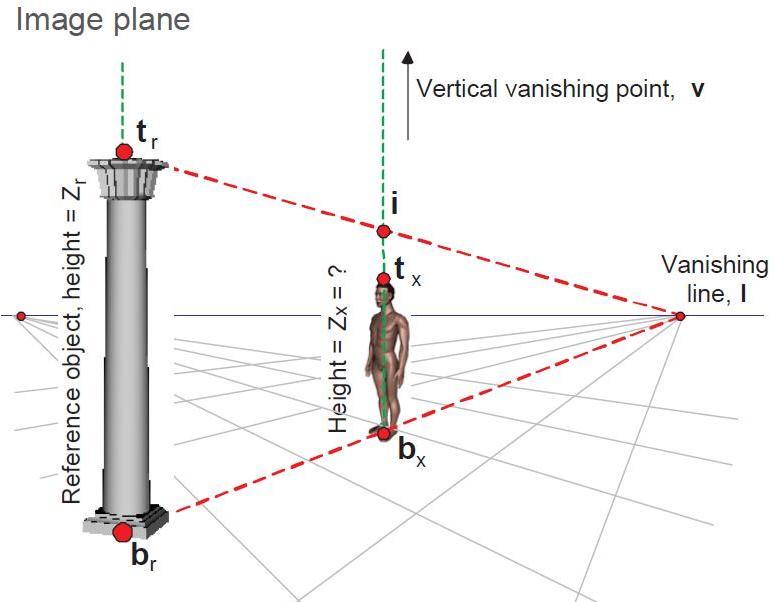}}
 \end{subfigure}
\caption{(a) Images of reference object used to compute homography. (b) SIFT correspondences betweem image of reference object and the scene. 
(c) Determine vanishing line \textbf{l} using 2 sets of parallel lines pair. 
(d) Determine vanishing point \textbf{v} using one set of parallel lines pair. 
(e) Illustration of height measurement using vanishing points and a reference height (image taken from \cite{MITguy}).}
\end{figure*} 

\section{Results}
\subsection{DimensionsApp}
Some salient features about DimensionsApp are: 
\begin{itemize}
\item Automatically finds vanishing line and vanishing point in the scene.
\item Automatically align user input points to the reference direction.
\item Can be used with almost any predefined reference object.
\item Instant accurate results. Error +/- 5 mm.
\item Small size, 2.4 MB and uses only 8.4 MB RAM.
\end{itemize}
\begin{figure*}[H]
\centering 
\begin{subfigure}[]{\includegraphics[width=0.32\linewidth]{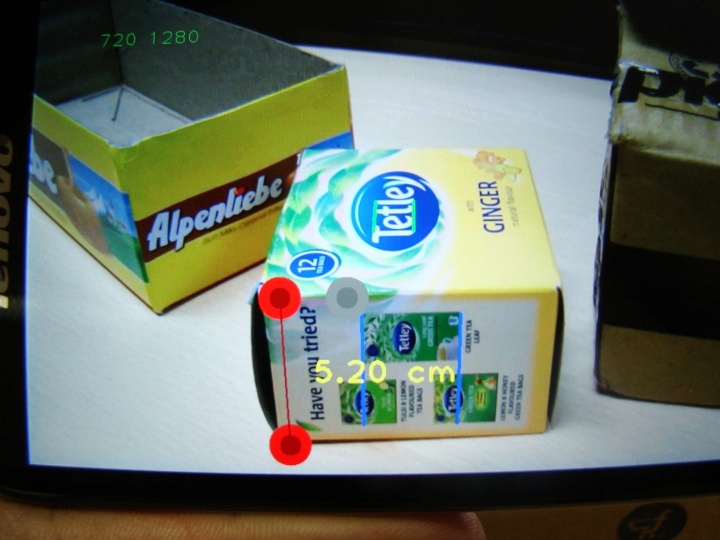}}
 \end{subfigure}    
 \begin{subfigure}[]{\includegraphics[width=0.32\linewidth]{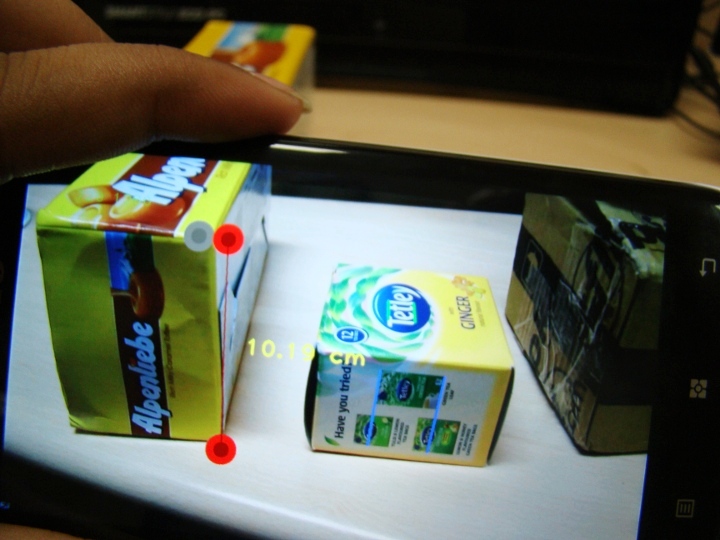}}
 \end{subfigure}    
 \begin{subfigure}[]{\includegraphics[width=0.32\linewidth]{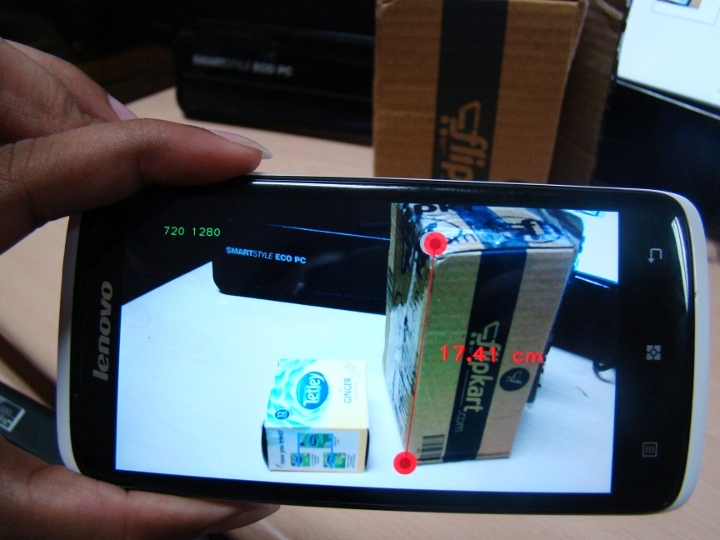}}
 \end{subfigure}    
\caption{Few examples of DimensionsApp. (a) Actual height = 5cm, estimated height= 5.2cm. (b) Actual height = 10cm, estimated height= 10.19cm. (c) Actual height = 17cm, estimated height= 17.41cm. All errors are within +/- 5mm.}
\end{figure*} 

\begin{figure}[b]
\centering 
\begin{subfigure}[]{\includegraphics[width=0.45\linewidth]{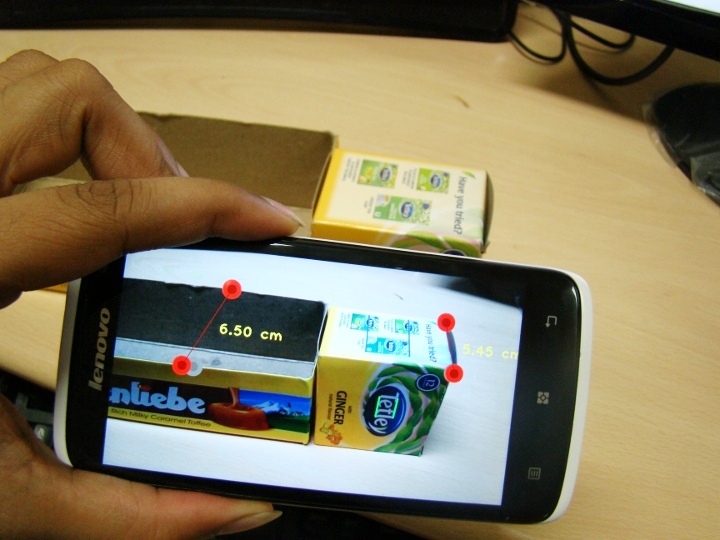}}
 \end{subfigure}    
 \begin{subfigure}[]{\includegraphics[width=0.45\linewidth]{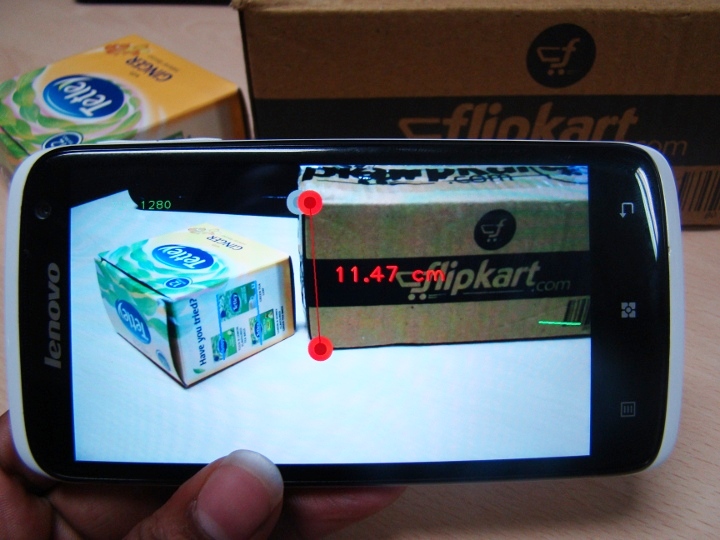}}
 \end{subfigure}    
\caption{Failure cases for DimensionsApp. (a) Actual Depth = 10cm, estimated height= 6.5cm. (b) Failure due to bad correspondences while computing homography.}
\end{figure}

{
\small
\bibliographystyle{ieee}
\bibliography{ego_activity_cvpr2015bib}

\begin{thebibliography}{1}\itemsep=-1pt

\bibitem{Criminisi}
I.~R. A.~Criminisi and A.~Zisserman.
\newblock Single view metrology.
\newblock In {\em IJCV}, 2000.

\bibitem{Reid}
I.~Reid and A.~Zisserman.
\newblock Goal-directed video metrology.
\newblock In {\em ECCV}, 1996.

\bibitem{MITguy}
S.-Y. Sun.
\newblock An implementation of single-view metrology.
\newblock In {\em Technical Report}, 2009.

\bibitem{Kim}
Y.~S. T.~Kim and K.~Hong.
\newblock Physics-based 3d position analysis of a soccer ball from monocular
  image sequences.
\newblock In {\em ICCV}, 1998.

\end{thebibliography}
}

\end{document}